\documentclass[10pt]{article}

\usepackage{graphicx}
\usepackage{authordate1-4}
\begin{document}
\title{Computational models: Bottom-up and top-down aspects}
\author{Laurent Itti and Ali Borji, University of Southern California}
\date{}
\maketitle

\section{Preliminaries and definitions}

Computational models of visual attention have become popular over the past decade, we believe primarily for two reasons:
First, models make testable predictions that can be explored by experimentalists as well as theoreticians; second,
models have practical and technological applications of interest to the applied science and engineering communities. In
this chapter, we take a critical look at recent attention modeling efforts.  We focus on {\em computational models of
  attention} as defined by Tsotsos \& Rothenstein \shortcite{Tsotsos_Rothenstein11}: Models which can process any visual
stimulus (typically, an image or video clip), which can possibly also be given some task definition, and which make
predictions that can be compared to human or animal behavioral or physiological responses elicited by the same stimulus
and task. Thus, we here place less emphasis on abstract models, phenomenological models, purely data-driven fitting or
extrapolation models, or models specifically designed for a single task or for a restricted class of stimuli. For
theoretical models, we refer the reader to a number of previous reviews that address attention theories and models more
generally
\cite{Itti_Koch01nrn,Paletta_etal05,Frintrop_etal10,Rothenstein_Tsotsos08,Gottlieb_Balan10,Toet11,Borji_Itti12pami}.

To frame our narrative, we embrace a number of notions that have been popularized in the field, even though many of them
are known to only represent coarse approximations to biophysical or psychological phenomena. These include the {\em
  attention spotlight} metaphor \cite{Crick84}, the role of focal attention in {\em binding} features into coherent
representations \cite{Treisman_Gelade80}, and the notions of an {\em attention bottleneck}, a {\em nexus}, and an {\em
  attention hand} as embodiments of the attentional selection process \cite{Rensink00,Navalpakkam_Itti05vr}. Further, we
cast the problem of modeling attention computationally as comprising at least three facets: {\em guidance} (that is,
which computations are involved in deciding where or what to attend to next?), {\em selection} (how is attended
information segregated out of other incoming sensory information?), and {\em enhancement} (how is the information
selected by attention processed differently than non-selected information?). While different theories and models have
addressed all three aspects, most computational models as defined above have focused on the initial and primordial
problem of guidance. Thus guidance is our primary focus, and we refer the reader to previous reviews on selection and
enhancement
\cite{allport1993attention,desimone1995neural,Reynolds_Desimone99,Driver01,robertson2003binding,Carrasco11}. Note that
guidance of attention is often thought of as involving {\em pre-attentive} computations to attract the focus of
attention to the next most behaviorally relevant location (hence, attention guidance models might~---~strictly
speaking~---~be considered pre-attention rather than attention models).

We explore models for exogenous (or bottom-up, stimulus-driven) attention guidance as well as for endogenous (or
top-down, context-driven, or goal-driven) attention guidance. Bottom-up models process sensory information primarily in
a feed-forward manner, typically applying successive transformations to visual features received over the entire visual
field, so as to highlight those locations which contain the most interesting, important, conspicuous, or so-called {\em
  salient} information \cite{Koch_Ullman85,Itti_Koch01nrn}. Many, but not all, of these bottom-up models embrace the
concept of a topographic saliency map, which is a spatial map where the map value at every location directly represents
visual salience, abstracted from the details of why a location is salient or not \cite{Koch_Ullman85}. Under the
saliency map hypothesis, the task of a computational model is then to transform an image into its spatially
corresponding saliency map, possibly also taking into account temporal relations between successive video frames of a
movie \cite{Itti_etal98pami}. Many models thus attempt to provide an operational definition of salience in terms of some
image transform or some importance operator that can be applied to an image and that directly returns salience at every
location, as we further examine below.

By far, the bottom-up, stimulus-driven models of attention have been more developed, probably because they are
task-free, and thus often require no learning, training, or tuning to open-ended task or contextual information. This
makes the definition of a purely bottom-up importance operator tractable.  Another attractive aspect of bottom-up
models~---~and especially saliency map models~---~is that, once implemented, they can easily be applied to any image and
yield some output that can be tested against human or animal experimental data.  Thus far, the most widely used test to
validate predictions of attention models has been direct comparison between model output and eye movements recorded from
humans or animals watching the same stimuli as given to the models (see, e.g., among many others,
\cite{Parkhurst_etal02,Itti06vc,le2006coherent}). Recently, several standard benchmark image and video datasets with
corresponding eye movement recordings from pools of observers have been adopted, which greatly facilitates quantitative
comparisons of computational models \cite{Carmi_Itti06jov,Judd_etal09,Toet11,Li_etal11,Borji_etal12tip}. It is important
to note, however, that several alternative measures have been employed as well (e.g., mouse clicks, search efficiency,
reaction time, optimal information gain, scanpath similarity; see \cite{Eckstein_etal09,Borji_Itti12pami}).  One caveat
of metrics that compare model predictions to eye movements is that the distinction between covert and overt attention is
seldom explicitly addressed by computational models: models usually produce as their final result a saliency map without
further concern of how such map may give rise to an eye movement scanpath \cite{Noton_Stark71} (but see
\cite{Itti_etal03spienn,Ma_Deng09,Sun_etal12} and active vision/robotics systems like
\cite{Orabona_etal05,Frintrop06,Belardinelli_etal06,ajallooeian2009fast}). Similarly, biases in datasets, models, and/or
behavior may affect the comparison results \cite{Tatler_Vincent09,tseng2009quantifying,Bonev_etal12}. While these are
important for specialist audiences, here we should just remember that quantitative model evaluation metrics based on eye
movements exist and are quite well established, although they remain approximate and should be used with caution
\cite{Tatler_etal05}.  Beyond metric issues, one last important consideration to ensure a valid direct comparison
between model and eye movements is that conditions should be exactly matched between the model run and the experimental
participants, which has not always been the case in previous work, as we discuss below.

While new bottom-up attention models are constantly proposed which rely on novel ways of analyzing images to determine
the most interesting or salient locations (we list over 50 of them below), several research efforts have also started to
address the more complex problem of building top-down models that are tractable and can be implemented
computationally. In the early days, top-down models have been mostly descriptive as they operated at the level of
conceptual entities (e.g., collections of objects present in a scene, to be evaluated in terms of a mental model of
which objects might be more important to a particular task of interest \cite{Ballard_etal95}), and hence have lacked
generalized computational implementations (because, for example, algorithms that can robustly recognize objects in
scenes were not available). As some of these hurdles have been alleviated by the advent of powerful new theories and
algorithms for object recognition, scene classification, decision making under uncertainty, machine learning, and formal
description languages that can capture task descriptions and background knowledge, exciting new models are beginning to
emerge which significantly surpass purely bottom-up approaches in their ability to predict attention behavior in complex
situations. Implementing complete, autonomous computational models of top-down attention is particularly important to
our theoretical understanding of task and context effects on attention, as these implementations remind us that some
assumptions often made to develop abstract models may not give rise to tractable computational models (e.g., it is easy
to note how humans are able to directly fixate a jar of jam when it is the next object required to make a sandwich
\cite{Land_Hayhoe01}~---~but how did they know that a jar is present and where it is located, if not through some
previous bottom-up analysis of the visual environment?).  As we further discuss in this chapter, these new computational
models also often blur the somewhat artificial dichotomy between bottom-up and top-down processing, since the so-called
top-down models do rely to a large extent onto a bottom-up flow of incoming information that is merged with goals and
task demands to give rise to the decision of where to attend next.

\begin{figure}[htb]
\includegraphics[width=\linewidth]{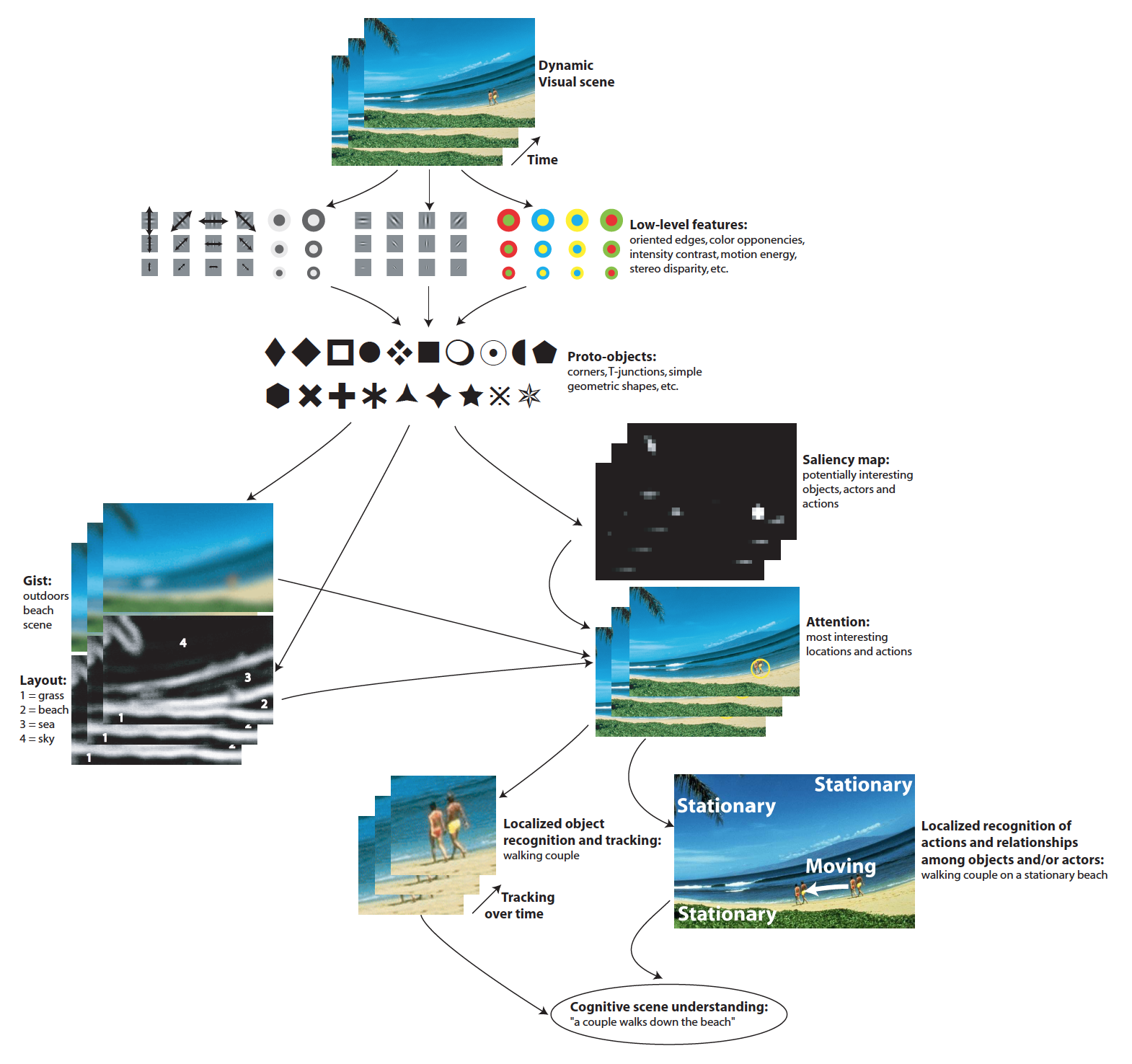}
\caption{Minimal attention-based architecture for complex dynamic visual scene understanding. This diagram augments the
  triadic architecture of Rensink (2000), which identified three key components of visual processing: volatile
  (instantaneous) and parallel pre-attentive processing over the entire visual field, from the lowest-level features up
  to slightly more complex proto-object representations (top), identification of the setting (scene gist and layout;
  left), and attentional vision including detailed and more persistent object recognition within the spatially
  circumscribed focus of attention (right). Here we have extended Rensink's architecture to include a saliency map to
  guide attention bottom-up towards salient image locations, and action recognition in dynamic scenes. Also see
  Navalpakkam~{\em et al.} (2005). }\label{FIGoverview}
\end{figure}

To frame the concepts exposed so far into a broader picture, we refer to Figure~\ref{FIGoverview} as a possible anchor
to help organize our thoughts and discussions of how different elements of a visual scene understanding system may work
together in the primate brain. In practice, few system-level efforts have included all components mentioned in
Figure~\ref{FIGoverview}, and most of our discussion will focus on computational models that implement parts of such a
system.

In what follows, we first examine in more details the key concepts of early bottom-up attention models
(Section~\ref{SECebu}), and then provide an overview and comparison of many subsequent models that have provided new
exciting insight into defining and computing bottom-up salience and attention (Section~\ref{SECbum}). We then turn to
top-down models (Section~\ref{SECtd}), first motivating them from experimental evidence, and then examining in turn
top-down models that modulate feature gains (Section~\ref{SECtdgains}), that derive spatial priors from the gist of the
visual scene (Section~\ref{SECtdspace}), and that implement more complex information foraging and decision making
schemes (Section~\ref{SECtdcomplex}). Finally, we discuss in Section~\ref{SECdisc} lessons learned from these models,
including on the nature and interaction between bottom-up and top-down processes, and promising directions towards the
creation of even more powerful combined bottom-up and top-down models.

\nocite{Navalpakkam_etal05noa,Rensink00}

\section{Early bottom-up attention concepts and models} \label{SECebu}

\begin{figure}
\includegraphics[width=\linewidth]{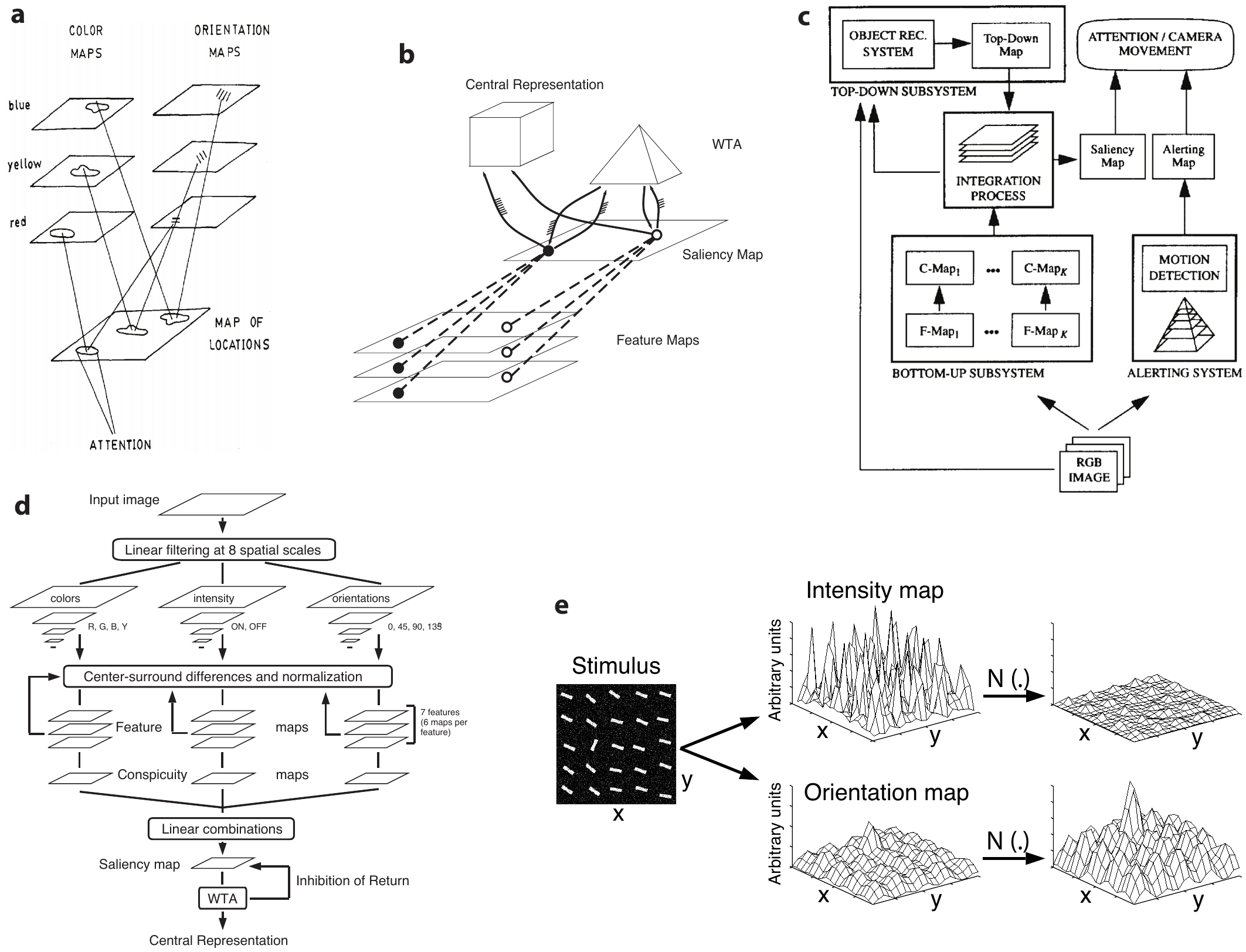}
\caption{Early bottom-up attention theories and models. {\bf (a)} Feature integration theory of Treisman \& Gelade
  (1980) posits several feature maps, and a focus of attention that scans a map of locations and collects and binds
  features at the currently attended location. (from Treisman \& Souther (1985)). {\bf (b)} Koch \& Ullman (1985)
  introduced the concept of a saliency map receiving bottom-up inputs from all feature maps, where a winner-take-all
  (WTA) network selects the most salient location for further processing. {\bf (c)} Milanese {\em et al.} (1994)
  provided one of the earliest computational models. They included many elements of the Koch \& Ullman framework, and
  added new components, such as an alerting subsystem (motion-based saliency map) and a top-down subsystem (which could
  modulate the saliency map based on memories of previously recognized objects). {\bf (d)} Itti {\em et al.}  (1998)
  proposed a complete computational implementation of a purely bottom-up and task-independent model based on Koch \&
  Ullman's theory, including multiscale feature maps, saliency map, winner-take-all, and inhibition of return. {\bf (e)}
  One of the key elements of Itti {\em et al.}'s model is to clearly define an attention interest operator, here denoted
  N(.), whereby the weight by which each feature map contributes to the final saliency map depends on how busy the
  feature map is. This embodies the idea that feature maps where one location significantly stands out from all others
  (as is the case in the orientation map shown) should strongly contribute to salience because they clearly vote for a
  particular location in space as the next focus of attention. In contrast, feature maps where many locations elicit
  comparable responses (e.g., intensity map shown) should not strongly contribute because they provide no clear
  indication of which location should be looked at next.} \label{FIGearlybu}
\end{figure}

Early attention models have been primarily influenced by the Feature Integration Theory \cite{Triesman_Gelade80},
according to which incoming visual information is first analyzed by early visual neurons which are sensitive to
elementary visual features of the stimulus (e.g., colors, orientations, etc). This analysis, operated in parallel over
the entire visual field and at multiple spatial and temporal scales, gives rise to a number of cortical feature maps,
where each map represents the amount of a given visual feature at any location in the visual field. Attention is then
the process by which a region in space is selected and features within that region are re-assembled or bound back
together to yield more complex object representations (Figure~\ref{FIGearlybu}.a). Koch and Ullman
\shortcite{Koch_Ullman85} extended the theory by advancing the concept of a single topographic and scalar {\em saliency
  map}, receiving inputs from the feature maps, as a computationally efficient representation upon which to operate the
selection of where to attend next: A simple maximum-detector or {\em winner-take-all} neural network
\cite{Arbib_Didday71} was proposed which would simply pick the next most salient location as the next attended one,
while an active {\em inhibition-of-return} \cite{Posner80} mechanism would later inhibit that location and thereby allow
attention to shift as the winner-take-all network would pick the next most salient location
(Figure~\ref{FIGearlybu}.b). From these ideas, a number of fully computational models started to be developed (e.g.,
Figure~\ref{FIGearlybu}.c,d).

At the core of these early models is the notion of visual salience, a signal that is computed in a stimulus-driven
manner and which indicates that some location is significantly different from its surroundings and is worthy of
attention.  Early models computed visual salience from bottom-up features in several feature maps, including luminance
contrast, red-green and blue-yellow color opponency, and oriented edges \cite{Itti_etal98pami}. While visual salience is
sometimes carelessly described as a physical property of a visual stimulus, it is important to remember that salience is
the consequence of an interaction of a stimulus with other stimuli, as well as with a visual system (biological or
artificial). For example, a color-blind person will have a dramatically different experience of visual salience than a
person with normal color vision, even when both look at exactly the same colorful physical scene.  Nevertheless, because
visual salience is believed to primarily arise from fairly low-level and stereotypical computations in the early stages
of visual processing, the factors contributing to salience are generally quite comparable from one observer to the next,
leading to similar experiences across a range of observers and of viewing conditions.

\nocite{Treisman_Gelade80,Treisman_Souther85,Milanese_etal94}

The essence of salience lies in enhancing the neural and perceptual representation of locations whose local visual
statistics significantly differ from the broad surrounding image statistics, in some behaviorally relevant manner. This
basic principle is intuitively motivated as follows. Imagine a simple search array as depicted in
Figure~\ref{FIGearlybu}.e, where one bar pops-out because of its unique orientation. Now imagine examining a feature map
which is tuned to stimulus intensity (luminance) contrast: because there are many white bars on a black background,
early visual neurons sensitive to local intensity contrast will respond vigorously to each of the bars (distractors and
target alike, since all have identical intensity). Based on the pattern of activity in this map, in which essentially
every bar elicits a strong peak of activity, one would be hard pressed to pick one location as being clearly more
interesting and worthy of attention than any of the others. Intuitively, hence, one might want to apply some
normalization operator {\cal N}(.) which would give a very low overall weight to this map's contribution to the final
saliency map. The situation is quite different when examining a feature map where neurons are tuned to local vertical
edges. In this map, one location (where the single roughly vertical bar is) would strongly excite the neural feature
detectors, while all other locations would elicit much weaker responses. Hence, one location clearly stands out and
hence becomes an obvious target for attention. It would be desirable in this situation that the normalization operator
{\cal N}(.) give a high weight to this map's contribution to the final saliency map
\cite{Itti_etal98pami,Itti_Koch00vr,Itti_Koch01nrn}.

Early bottom-up attention models have created substantial interest and excitement in the community, especially as they
were shown to be applicable to an unconstrained variety of stimuli, as opposed to more traditional computer vision
approaches at the time, which often had been designed to solve a specific task in a specific environment (e.g., detect
human faces in photographs taken from a standing human viewpoint \cite{Viola_Jones01}).  Indeed, no parameter tuning nor
any prior knowledge related to the contents of the images or video clips to be processed was necessary for any of many
early results, as the exact same model processed psychophysical stimuli, filmed outdoors scenes, Hollywood movie
footage, video games, and robotic imagery. This gave rise to model prediction results that included, for example, the
reproduction by Itti {\em et al.}'s model of human behavior in visual search tasks (e.g., pop-out versus conjunctive
search \cite{Itti_Koch00vr}); demonstration of strong robustness to image noise \cite{Itti_etal98pami}; automatic
detection of traffic signs and other salient objects in natural environments filmed by a consumer-grade color video
camera \cite{Itti_Koch01ei}; the detection of pedestrians in natural scenes \cite{Miau_etal01spie}; and of military
vehicles in overhead imagery \cite{Itti_etal01oe}; and~---~most importantly~---~the widely demonstrated ability of the
model to predict where humans look when freely images or videos that range from search arrays to fractals to satellite
images to everyday indoors and outdoors scenes \cite{Parkhurst_etal02,Peters_etal05vr}.

\section{Flourishing of bottom-up models} \label{SECbum}

\begin{figure}[htb]
\includegraphics[width=\linewidth]{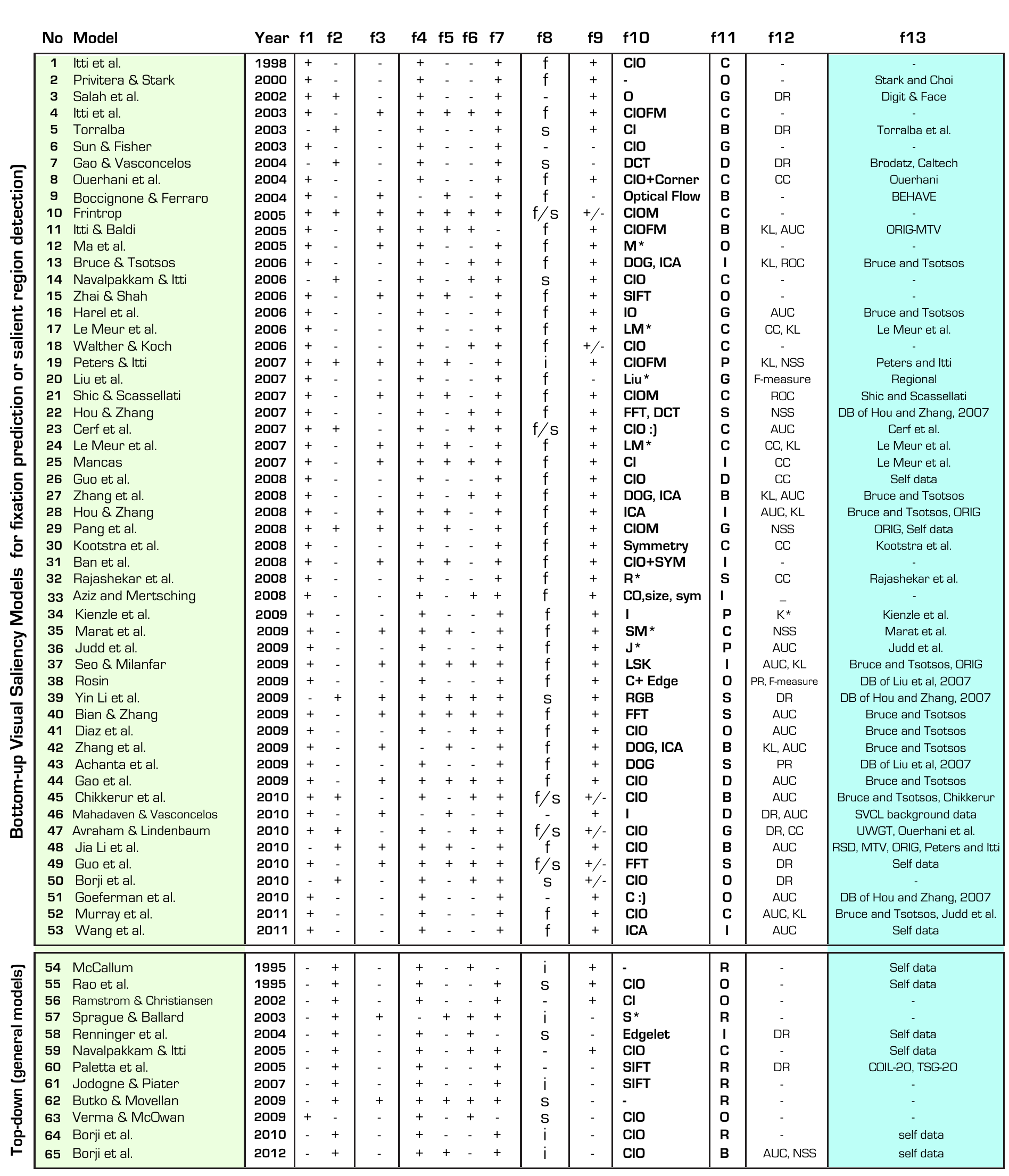}
\caption{\footnotesize Survey of bottom-up and top-down computational models, classified according to 13
  factors. Factors in order are: Bottom-up ($f_{1}$), Top-down ($f_{2}$), Spatial (-)/Spatio-temporal (+) ($f_{3}$),
  Static ($f_{4}$), Dynamic ($f_{5}$), Synthetic ($f_{6}$) and Natural ($f_{7}$) stimuli, Task-type ($f_{8}$),
  Space-based(+)/Object-based(-) ($f_{9}$), Features ($f_{10}$), Model type ($f_{11}$), Measures ($f_{12}$), and Used
  dataset ($f_{13}$). In Task type ($f_{8}$) column: free-viewing ($f$); target search ($s$); interactive ($i$). In
  Features ($f_{10}$) column: CIO: color, intensity and orientation saliency; CIOFM: CIO plus flicker and motion
  saliency; M* = motion saliency, static saliency, camera motion, object (face) and aural saliency (Speech-music); LM* =
  contrast sensitivity, perceptual decomposition, visual masking and center-surround interactions; Liu* =
  center-surround histogram, multi-scale contrast and color spatial-distribution; R* = luminance, contrast,
  luminance-bandpass, contrast-bandpass; SM* = orientation and motion; J* = CIO, horizontal line, face, people detector,
  gist, etc; S* = color matching, depth and lines; :) = face. In Model type ($f_{11}$) column, R means that a model is
  based RL. In Measures ($f_{12}$) column: K* = used Wilcoxon-Mann-Whitney test (The probability that a random chosen
  target patch receives higher saliency than a randomly chosen negative one); DR means that models have used a measure
  of detection/classification rate to determine how successful was a model. PR stands for Precision-Recall. In dataset
  ($f_{13}$) column: Self data means that authors gathered their own data. For a detailed definition of these factors
  please refer to Borji \& Itti (2012 PAMI).}
\label{FIGmodels}
\end{figure}

Following initial success, many research groups started exploring the notions of bottom-up attention and visual
salience, which has given rise to many new computational models. We summarize 53 bottom-up models along 13 different
factors in Figure~\ref{FIGmodels}. A thorough examination of all models is certainly not feasible in this limited
space. Instead, we highlight below the main trends in seven categories that span from strong inspiration from biological
vision to more abstract mathematical definitions and implementations of the concept of saliency. Models can coarsely be
categorized as follows (but also see \cite{Tsotsos_Rothenstein11} for another possible taxonomy).  Please note that some
models fall under more than one category.

\textbf{Cognitive models}. Research into saliency modeling escalated after Itti {\em et
  al.}'s~\shortcite{Itti_etal98pami} implementation of Koch and Ullman's~\shortcite{Koch_Ullman85} computational
architecture based on the Feature Integration Theory \cite{Triesman_Gelade80}. In cognitive models, which were the first
ones to approach the problem of algorithmically computing saliency in arbitrary digital images, an input image is
decomposed into a set of feature maps across spatial scales which are then linearly or non-linearly normalized and
combined to form a master saliency map.  An important element of this theory is the idea of center-surround which
defines saliency as distinctiveness of an image region to its immediate surroundings.  Almost all saliency models are
directly or indirectly inspired by cognitive concepts of visual attention (e.g.,~\cite{le2006coherent,MaratIJCV}).

\textbf{Information-theoretic models}. Stepping back from biological implementation machinery, models in this category
are based on the premise that localized saliency computations serve to maximize information sampled from one's
environment. These models assign higher saliency values to scene regions with rare features.  Information of visual
feature $F$ is $I(F)= - log \mbox{ $p(F)$}$ which is inversely proportional to the likelihood of observing $F$ (i.e.,
$p(F)$).  By fitting a distribution $P(F)$ to features (e.g., using Gaussian Mixture Model or Kernels), rare features
can be immediately found by computing $P(F)^{-1}$ in an image.  While in theory using any feature space is feasible,
usually these models (inspired by efficient coding representations in visual cortex) utilize a sparse set of basis
functions (using ICA filters) learned from a repository of natural scenes. Some basic approaches in this domain are
AIM~\cite{BruceNIPS}, Rarity~\cite{mancas2007computational}, LG (Local + Global image patch
rarity)~\cite{Borji_Itti12cvpr}, and incremental coding length models~\cite{HouZhangNIPS2008}.

\textbf{Graphical models}. Graphical models are generalized Bayesian models which have been employed for modeling
complex attention mechanisms over space and time. Torralba~\cite{Torralba03} proposed a Bayesian approach for modeling
contextual effects on visual search which was later adopted in the SUN model~\cite{ZhangTong2Etal008} for fixation
prediction in free viewing. Itti and Baldi~\cite{Itti_Baldi05cvpr} defined surprising stimuli as those which
significantly change beliefs of an observer. Harel {\em et al.} (GBVS)~\cite{harel2007graph} propagated similarity of
features in a fully connected graph to build a saliency map.  Avraham~\cite{Avraham_Lindenbaum10}, Jia Li~{\em et
  al.},~\cite{li2010optimol}, and Tavakoli~{\em et al.}~\cite{rezazadegan2011fast}, have also exploited Bayesian
concepts for saliency modeling.

\textbf{Decision theoretic models}. This interpretation states that attention is driven optimally with respect to the
end task. Gao and Vasconcelos~\cite{gao2004discriminant} argued that for recognition, salient features are those that
best distinguish a class of objects of interest from all other classes. Given some set of features $X = \{X_{1},
\cdots,X_{d}\}$, at locations $l$, where each location is assigned a class label $Y$ with $Y_{l} = 0$ corresponding to
background and $Y_{l} = 1$ indicates objects of interest, saliency is then a measure of mutual information (usually
Kullback-Leibler divergence (KL)), computed as $I(X,Y) = \sum_{i=1}^{d} I(X_{i}, Y)$. Besides having good accuracy in
predicting eye fixations, these models have been very successful in computer vision applications (e.g., anomaly
detection and object tracking).

\textbf{Spectral-analysis models}. Instead of processing an image in the spatial domain, these models derive saliency in
the frequency domain. This way, there is no need for image processing operations such as center-surround or
segmentation.  Hou and Zhang~\cite{HouZhangCVPR2007} derive saliency for an image with amplitude $\mathcal{A}(f)$ and
phase $\mathcal{P}(f)$ as follows. The log spectrum $\mathcal{L}(f)$ is computed from the down-sampled image. From
$\mathcal{L}(f)$, the spectral residual $\mathcal{R}(f)$ is obtained by multiplying $\mathcal{L}(f)$ with $h_{n}(f)$
which is an $n \times n$ local average filter and subtracting the result from itself. Saliency map is then the inverse
Fourier transform of the exponential of amplitude plus phase (i.e., $\mathcal{S}(x) = \mathcal{F}^{-1}\big[exp
  \big(\mathcal{R}(f) +\mathcal{P}(f) \big)\big]$).  The saliency of each point is squared to indicate the estimation
error and is then smoothed with a Gaussian filter for better visual effect.  Bian and Zhang~\cite{BianZhang2009} and Guo
{\em et al.}~\cite{GuoIEEEIP} proposed spatio-temporal models in the spectral domain.

\textbf{Pattern classification models}. Models in this category use machine learning techniques to learn
``stimuli-saliency'' mappings from image features to eye fixations. They estimate saliency $s$; $p(s|f)$ where $f$ is a
feature vector which could be the contrast of a location and its surrounding neighborhood. Kienzle~{\em et
  al.}~\cite{kienzle2007nonparametric}, Peters and Itti~\cite{Peters_Itti07cvpr}, and Judd~{\em et
  al.}~\shortcite{Judd_etal09} used image patches, scene gist, and a vector of several features at each pixel,
respectively and used classical SVM and Regression classifiers for learning saliency. In an extension of Judd model,
Borji \shortcite{Borji12cvpr} showed that using a richer set of features including bottom-up saliency maps of other
models and within-object regions (e.g., eye within faces) along with a boosting classifier leads to higher fixation
predicting accuracy. Tavakoli~{\em et al.}~\cite{rezazadegan2011fast}, used sparse sampling and kernel density
estimation to estimate the above probability in a Bayesian framework.  Note that some of these models may not be purely
bottom-up since they use features that guide top-down attention, for example faces or
text~\cite{Judd_etal09,Cerf_etal08}.

\textbf{Other models}. Some other models exist that do not easily fit into our categorization. For example, Seo and
Milanfar \cite{SeoMilanfar2009JV} proposed self-resemblance of local image structure for saliency detection. The idea of
decorrelation of neural response was used for a normalization scheme in the Adaptive Whitening Saliency (AWS)
model~\cite{garcia2009decorrelation}. Kootstra~{\em et al.}~\cite{kootstra2008paying} developed symmetry operators for
measuring saliency and Goferman~{\em et al.}  \cite{goferman2010context} proposed a context-aware saliency detection
model with successful applications in re-targeting and summarization.

An important trend to consider is that over the past years starting from~\cite{LiuSunZhengTangShumCVPR2007}, models have
begun to diverge into two different classes: models of \textit{fixation prediction} and models of \textit{salient
  region detection}. While the goal of the former models is to predict locations that grab attention, the latter models
attempt to segment the most salient object or region in a scene. A saliency operator is usually used to estimate the
extent of the object that is predicted to be the most likely first attended object. Evaluation is often done by
measuring precision-recall of saliency maps of a model against ground-truth data (explicit saliency judgments of
subjects by annotating salient objects or clicking on locations). Some models in two categories have compared themselves
against each other, without being aware of the distinction.

Figure~\ref{FIGmodels} shows a list of models and their properties according to thirteen qualitative criteria derived
from behavioral and computational studies. The majority (53 out of 65) of covered attention models consists of bottom-up
models, indicating that at least from a computational perspective it is easier to formulate attention guidance
mechanisms based on low-level image features. This is reinforced by the existence of several established benchmark
datasets and standard evaluation scores for bottom-up models. The situation is the opposite for top-down attention
modeling although we have recently initiated an effort to share data and code~\cite{Borji_etal12aaai,Borji_etal12smc}.

\begin{figure}[htb]
\includegraphics[width=\linewidth]{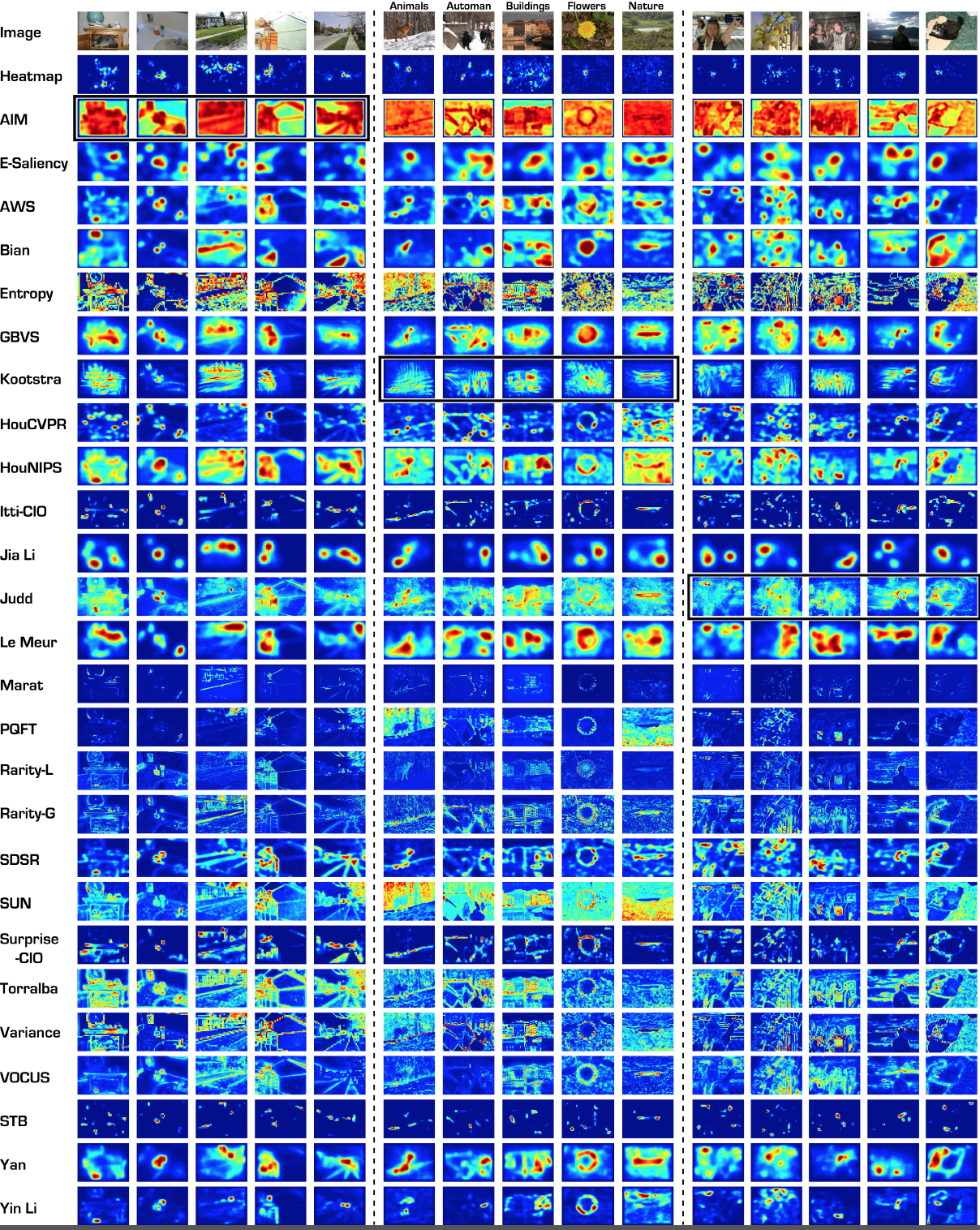}
\caption{Example images (first row), human eye movement heatmaps (second row), and saliency maps from 26 computational
  models. The three vertical dashed lines separate the three datasets used (Bruce \& Tsotsos, Kootstra \& Schomacker,
  and Judd {\em et al.}). Black rectangles indicate the model originally associated with a given image dataset. Please
  see Borji {\em et al.} (2012 TIP) for additional details.}
\label{FIGmaps}
\end{figure}

\begin{figure}[htb]
\includegraphics[width=3in]{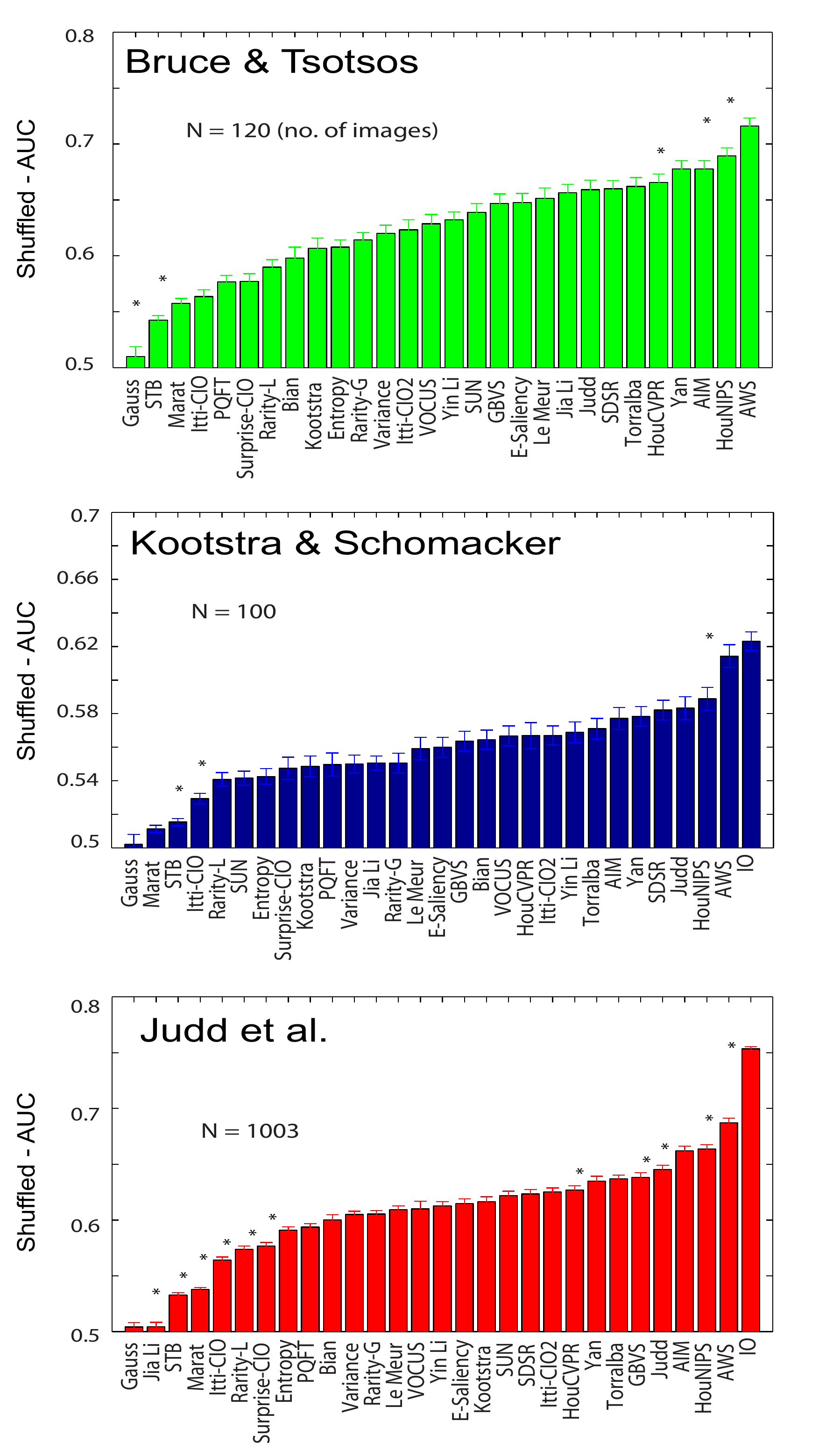}
\caption{Ranking visual saliency models over three image datasets. Left column: Bruce \& Tsotsos (2005), Middle column:
  Kootstra \& Shomaker (2008), and Right column: Judd {\em et al.} (2009) using shuffled AUC score. Stars indicate
  statistical significance using t-test (95\%, $p \le 0.05 $) between consecutive models. Error bars indicate standard
  error of the mean (SEM): $\frac{\sigma}{\sqrt{N}}$, where $\sigma$ is the standard deviation and $N$ is the number of
  images. The Judd model uses center feature, gist and horizon line, and object detectors for cars, faces, and human
  body. Itti-CIO2 is the approach proposed by Itti {\em et al.} (1998) that uses a normalization scheme known as
  Maxnorm: For each feature map, find the global max $M$ and find the average $m$ of all other local maxima. Then just
  weight the map by $(M-m)^2$. In the Itti-CIO method (Itti \& Koch, 2000), normalization is: Convolve each map by a
  Difference of Gaussian(DoG) filter, cut off negative values, and iterate this process for a few times. As results show
  the Maxnorm normalization scheme performs better. In the literature, majority of models have been compared against
  Itti-CIO model. Please see Borji {\em et al.} (2012 TIP) for additional details on these results.}
\label{FIGauc}
\end{figure}

A brief comparison of saliency maps of 26 models on a few test images (Figure~\ref{FIGmaps}) shows large differences in
appearance of the maps generated by different models. Some models generate very sparse maps while others are
smoother. This makes fair model comparison a challenge since some scores may be influenced by smoothness of a map
\cite{Tatler_etal05}.  Recently, Borji {\em et al.}  \shortcite{Borji_etal12tip} performed a detailed investigation of
models to quantify their correlations with human attentional behavior. This study suggests that so far the so-called
``shuffled AUC (Area Under the ROC Curve)'' score \cite{ZhangTong2Etal008} is the most robust (this score uses
distributions of human fixations on other stimuli than the one being scored to establish a baseline, which attenuates
the effects of certain biases in eye movements datasets, the strongest being a bias towards looking preferentially near
the center of any image). Results are shown in Figure~\ref{FIGauc}. This model evaluation shows a gap between current
models and human performance. This gap is smaller for some datasets, but overall exists. Discovering and adding more
top-down features to models will hopefully boost their performance. The analysis also shows that some models are very
effective (e.g., HouNIPS, Bian, HouCVPR, Torralba, and Itti-CIO2 in Figure~\ref{FIGauc}) and also very fast providing a
trade-off between accuracy and speed necessary for many applications.

Despite past progress in bottom-up saliency modeling and fixation prediction while freely viewing natural scenes,
several open questions remain that should be answered in the future. The most confusing one is that of ``center bias,''
whereby humans often appear to preferentially look near an image's center. It is believed to be largely caused by
stimulus bias (e.g., photographer bias, whereby photographers tend to frame interesting objects near the image center).
Collecting fixation datasets with no or less center bias, and studying its role on model evaluation needs to be
addressed with natural scenes (see, e.g., \cite{Parkhurst_etal02,Peters_etal05vr} for unbiased artificial datasets of
fractal images).  As opposed to saliency modeling on static scenes, the domain of spatio-temporal attention remains less
explored (See~\cite{Dorr_etal10,Wang_etal12} for examples).  Emphasis should be on finding cognitive factors (e.g.,
actor, non-actor) rather than simple bottom-up features (e.g., motion, flicker, or focus of expansion), and some of the
top-down models discussed below have started to explore how more semantic scene analysis can influence attention.
Another aspect is the study of attention on affective and emotional stimuli. Although a database of fixations on
emotional images has been gathered by Ramanathan~{\em et al.} \shortcite{Ramanathan_etal10}, it is still not clear
whether current models can be extended to explain such fixations.

\nocite{Borji_Itti12pami,Borji_etal12tip,kootstra2008paying}

\nocite{salah2002selective,gao2004discriminant,ouerhani2003real,boccignone2004modelling,ma2005generic,zhai2006visual,harel2007graph,le2006coherent,walther2006modeling,shic2007behavioral,mancas2007computational,guo2008spatio,pang2008stochastic,kootstra2008paying,ban2008dynamic,rajashekar2008gaffe,aziz2008fast,kienzle2007nonparametric,rosin2009simple,li2010visual,garcia2009decorrelation,zhang2009sunday,achanta2009frequency,gao2009discriminant,mahadevan2010spatiotemporal,li2010optimol,murray2011saliency,mccallum1996reinforcement,rao1996modeling,ramstrom2002visual,renninger2005information,jodogne2007closed,butko2009optimal,verma2009generating,borji2010online}

\nocite{BruceNIPS,kootstra2008paying,Judd_etal09}

\section{Top-down guidance of attention by task demands} \label{SECtd}

Research towards understanding the mechanisms of top-down attention has given rise to two broad classes of models:
models which operate on semantic content, and models which operate on raw pixels and images. Models in the first
category are not fully computational in the sense used in the present chapter, in that they require that an external
expert (typically, one or more humans) first pre-processes raw experimental recordings, often to create semantic
annotations (e.g., translate from recorded video frames and gaze positions into sequences that describe which objects
were being looked at).  For example, in a block copying task \cite{Ballard_etal95}, the observers' algorithm for
completing the task was revealed by their pattern of eye movements: first select a target block in the model by fixating
it, then find a matching block in the resource pool, then revisit the model to verify the block's position, then fixate
the workspace to place the new block in the corresponding position. Other studies have used naturalistic interactive or
immersive environments to give high-level accounts of gaze behavior in terms of objects, agents, ``gist'' of the scene,
and short-term memory \cite{Yarbus67,Henderson_Hollingworth99,Rensink00,Land_Hayhoe01,sodhi2002road,hayhoe2003visual},
to describe, for example, how task-relevant information guides eye movements while subjects make a sandwich
\cite{Land_Hayhoe01,hayhoe2003visual} or how distractions such as setting the radio or answering a phone affect eye
movements while driving \cite{sodhi2002road}.

While such perceptual studies have provided important constraints regarding goal-oriented high-level vision, additional
work is needed to translate these descriptive results into fully-automated computational models that can be used in the
application domains mentioned above. That is, although the block copying task reveals observers' algorithm for
completing the task, it does so only in the high-level language of ``workspace'' and ``blocks'' and ``matching.''  In
order for a machine vision system to replicate human observers' ability to understand, locate, and exploit such visual
concepts, we need a ``compiler'' to translate such high-level language into the assembly language of vision---that is,
low-level computations on a time-varying array of raw pixels. Unfortunately, a general computational solution to this
task is tantamount to solving computer vision.

\begin{figure}[htb]
\includegraphics[width=\linewidth]{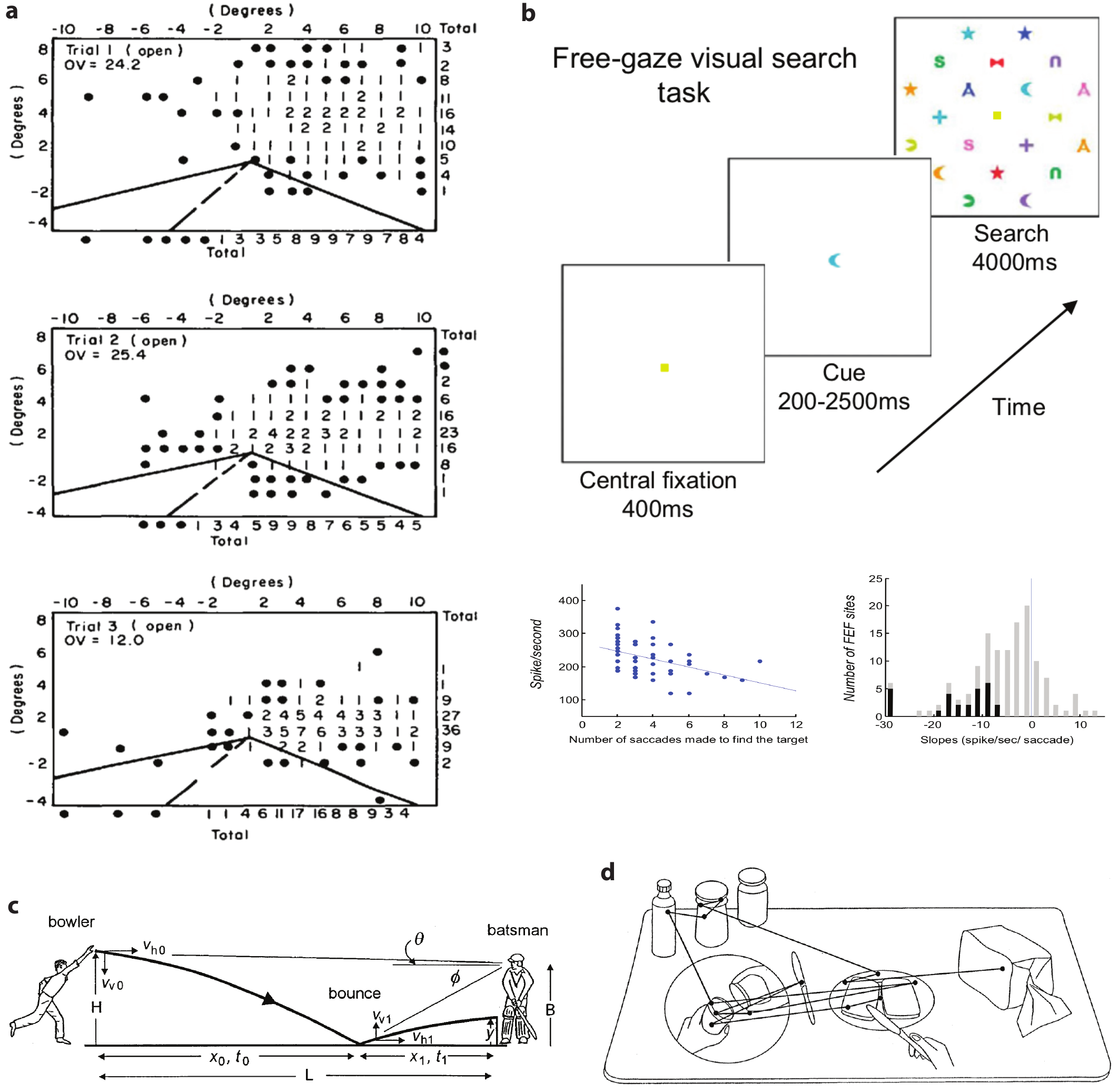}
\caption{Experimental motivation to explore spatial biases, feature biases, and more complex semantic top-down
  models. {\bf (a)} Percentage of fixations onto different road locations while human drivers drove at 50 miles/hour on
  an open road three times (once each panel; dots indicate non-zero percentages smaller than 1). We can see that over
  the three repetitions (top to bottom panels), eye fixations started clustering more tightly around the left side of
  the horizon line (from Mourant \& Rockwell, 1970). This motivates top-down models to learn over time how a task may
  induce some spatial biases in the deployment of attention. {\bf (b)} Neural recordings in the frontal eye fields (FEF)
  as monkeys searched from a cued target among various distractors reveals that the number of saccades the animal made
  to find the target is negatively correlated with neural firing at the target location around ($\pm$50ms) the onset of
  the first saccade (scatter plot shows examples, one dot per trial, from one recording site, and the negatively-sloped
  linear regression; histogram shows distributions of slopes over all recording sites, with the significant ones in
  black and all others in gray). This suggests that top-down models can also exploit biasing for specific features of a
  search target to attempt to guide attention faster towards the target (from Zhou \& Desimone, 2011). {\bf (c)} Eye
  movement recordings of cricket batsmen revealed that their ``eye movements monitor the moment when the ball is
  released, make a predictive saccade to the place where they expect it to hit the ground, wait for it to bounce, and
  follow its trajectory for 100-200 ms after the bounce'' (Land \& McLeod, 2000). This suggests that some knowledge of
  physics, gravity, bouncing, etc.\ may be necessary to fully understand human gaze behavior in this more complex
  scenario. {\bf (d)} Eye movement recordings while making a sandwich are clearly aimed towards the next required item
  during the unfolding of the successive steps required by the task, with very little searching or exploration (Land \&
  Hayhoe, 2001). Thus, recognition and memorization of objects in the scene is also likely to be required of top-down
  models to tackle such more complex scenario.}
\label{FIGtd}
\end{figure}

From behavioral and in particular eye-tracking experiments during execution of real-world tasks, several key
computational factors can be identified which can be implemented in computational models (Figure~\ref{FIGtd}):
\begin{itemize}
\item {\bf Spatial biases}, whereby a given high-level task or top-down set may make some region of space more likely to
  contain relevant information. For example, when the task is to drive, it is important to keep our eyes on the road
  (Figure~\ref{FIGtd}.a). We describe below how bottom-up attention models can be enhanced by considering such
  task-driven spatial constraints, for example to suppress salient stimuli that lie outside the task-relevant region of
  visual space. These models are motivated by both psychophysical and physiological evidence of spatial biasing of
  attention based on both short-term and long-term top-down cues \cite{Chun_Jiang98,Summerfield_etal10}, resulting in
  enhancement of attended visual regions and suppression of the un-attended ones
  \cite{Brefczynski_DeYoe99,Kastner_etal99};
\item {\bf Feature biases}, whereby the task may dictate that some visual features (e.g., some colors) are more likely
  associated with items of interest than other features (Figure~\ref{FIGtd}.b). Bottom-up models can also be enhanced to
  account for feature biases, for example by modulating according to top-down goals the relative weights by which
  different feature maps contribute to a saliency map (e.g., when searching for a blue item, increase the gain of
  blue-selective feature maps). These models also are motivated by experimental studies of so-called feature-based
  attention \cite{Treue_MartinezTrujillo99,Saenz_etal02,Zhou_Desimone11,MartinezTrujillo11} and, in particular, recent
  theories and experiments investigating the role of the pulvinar nucleus in carrying out such biases
  \cite{Baluch_Itti11tins,Saalmann_etal12};
\item {\bf Object-based and cognitive biases}, whereby knowing about objects, about how they may interact with each
  other, and about how they obey the laws of physics such as gravity and friction, may help humans make more efficient
  decisions of where to attend next to achieve a certain top-down goal (e.g., playing cricket, Figure~\ref{FIGtd}.c, or
  making a sandwich, Figure~\ref{FIGtd}.d). Models can be taught how to recognize objects and possibly other aspects of
  the world, to enable semantic reasoning that may give rise to these more complex top-down attention behaviors. These
  models are also motivated by recent experimental findings \cite{Vo_Henderson09,Schmidt_Zelinsky09,Hwang_etal11}.
\end{itemize}

We review top-down models that have implemented these strategies below. Although spatial biases have historically been
studied first, we start with feature biasing models as those are conceptually simpler extensions to the bottom-up
models described in the previous sections.

\nocite{Mourant_Rockwell70,Zhou_Desimone11,Land_McLeod00,Land_Hayhoe01}

\subsection{Top-down biasing of bottom-up feature gains} \label{SECtdgains}

A simple strategy to include top-down influences in a computational attention model is to modulate the low-levels of
visual processing of the model according to the top-down task demands. This embodies the concept of {\em feature-based
  attention}, whereby increased neural response can be detected in monkeys and humans to visual locations which contain
features that match a feature of current behavioral interest (e.g., locations that contain upward moving dots when the
animal's task is to monitor upward motion \cite{Treue_MartinezTrujillo99,Saenz_etal02,Zhou_Desimone11}.

\begin{figure}[htb]
\includegraphics[width=\linewidth]{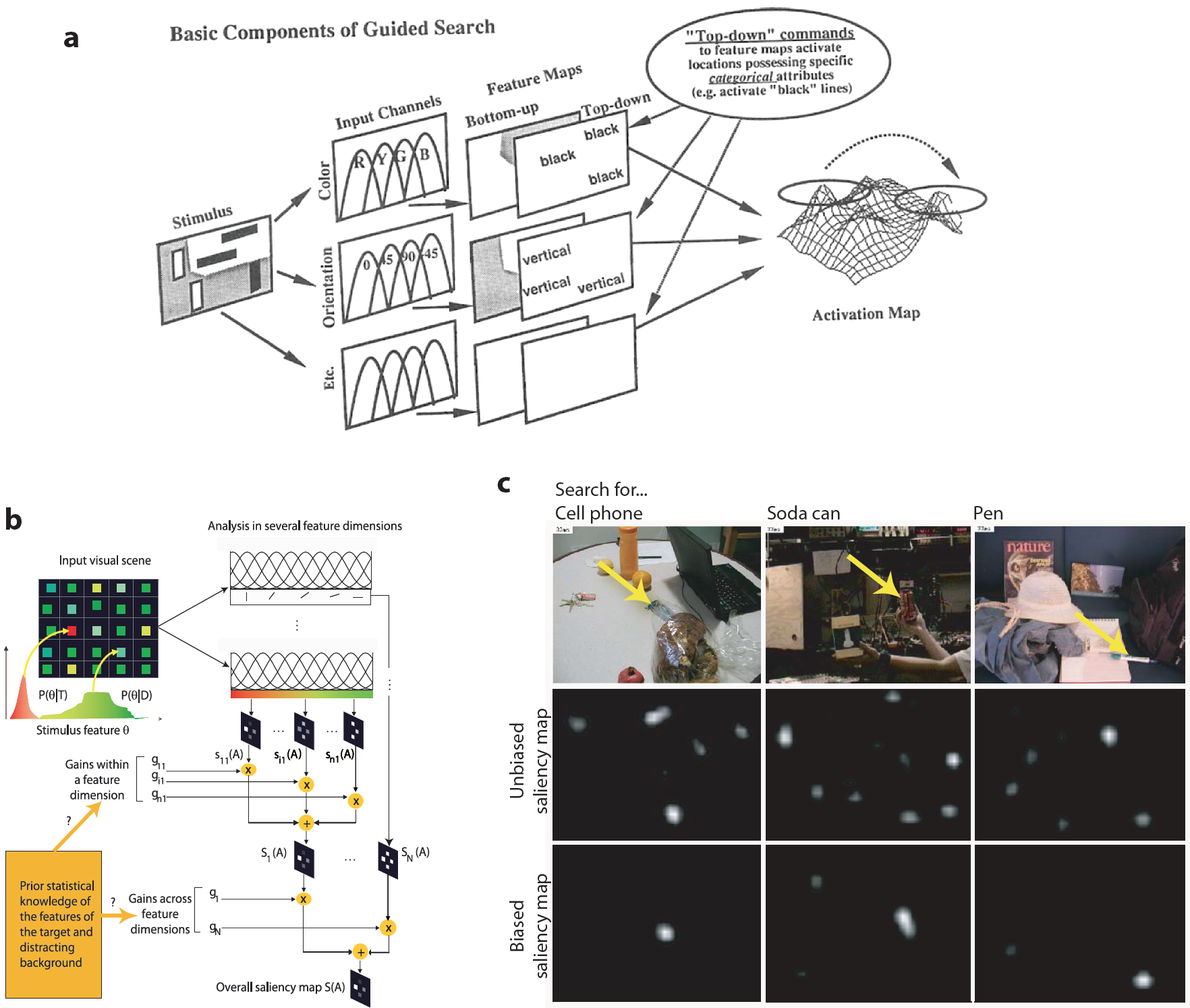}
\caption{Top-down models that modulate feature gains. {\bf (a)} The Guided Search theory of Wolfe (1994) predicted that
  bottom-up feature maps can be weighted and modulated by top-down commands. {\bf (b)} Computational framework to
  optimally compute the weight of each feature map, given top-down knowledge of the expected distribution of feature
  values $\theta$ for both target objects ($P(\theta|T)$) and distractors ($P(\theta|D)$). The gain of each feature is
  set inversely proportionally to the expected target-to-distractor signal-to-noise ratio given these distributions
  (Navalpakkam \& Itti, 2006; 2007). {\bf (c)} Examples of images (top row), naive un-weighted saliency maps (middle
  row), and optimally-biased saliency maps based on feature distributions gathered from sample training images (bottom
  row). Although the object of interest was not the most salient according to a purely bottom-up model in these
  examples, it becomes the most salient once the model is biased using top-down gains computed from the target and
  distractor feature distributions (Navalpakkam \& Itti, 2006).}
\label{FIGtdgains}
\end{figure}

While the idea of top-down feature biases was already present in early conceptual models like the {\em Guided Search}
theory \cite{Wolfe94} and {\em FeatureGate} \cite{Cave99}, the question for computational modelers has been how exactly
the feature gains should be adjusted to yield optimal expected enhancement of a desired target among unwanted
distractors (Figure~\ref{FIGtdgains}).  Earlier models have used supervised learning techniques to compute feature gains
from example images where targets of interest had been manually indicated
\cite{Itti_Koch01ei,Frintrop_etal05,borji2011cost}. A more recent approach uses eye movement recordings to determine
these weights \cite{Zhao_Koch11}. Interestingly, it has recently been proposed that an optimal set of weight can be
computed in closed-form given distributions of expected features for both targets of interest to the task and irrelevant
clutter or distractors. In this approach, each feature map is characterized by a target-to-distractor response ratio (or
signal-to-noise ratio, SNR), and feature maps are simply assigned a weight that is inversely proportional to their SNR
\cite{Navalpakkam_Itti06cvpr,Navalpakkam_Itti07n}. In addition to giving rise to a fully computational model, this
theory has also been found to explain many aspects of human guided search behavior
\cite{Navalpakkam_Itti07n,Serences_Saproo10}

Note how the models of Figure~\ref{FIGtdgains} start introducing a blur between the notions of bottom-up and top-down
processing. In these models, indeed, the effect of top-down knowledge is to modulate the way in which bottom-up
computations are carried out. This opens the question of whether a pure bottom-up state may ever exist, and what the
corresponding gain values may be (e.g., unity as has been assumed in many models?). These questions are also being
raised by a number of recent experiments \cite{Theeuwes10,Awh_etal12}, and are further discussed below.

\subsection{Spatial priors and scene context} \label{SECtdspace}

\begin{figure}[htb]
\includegraphics[width=\linewidth]{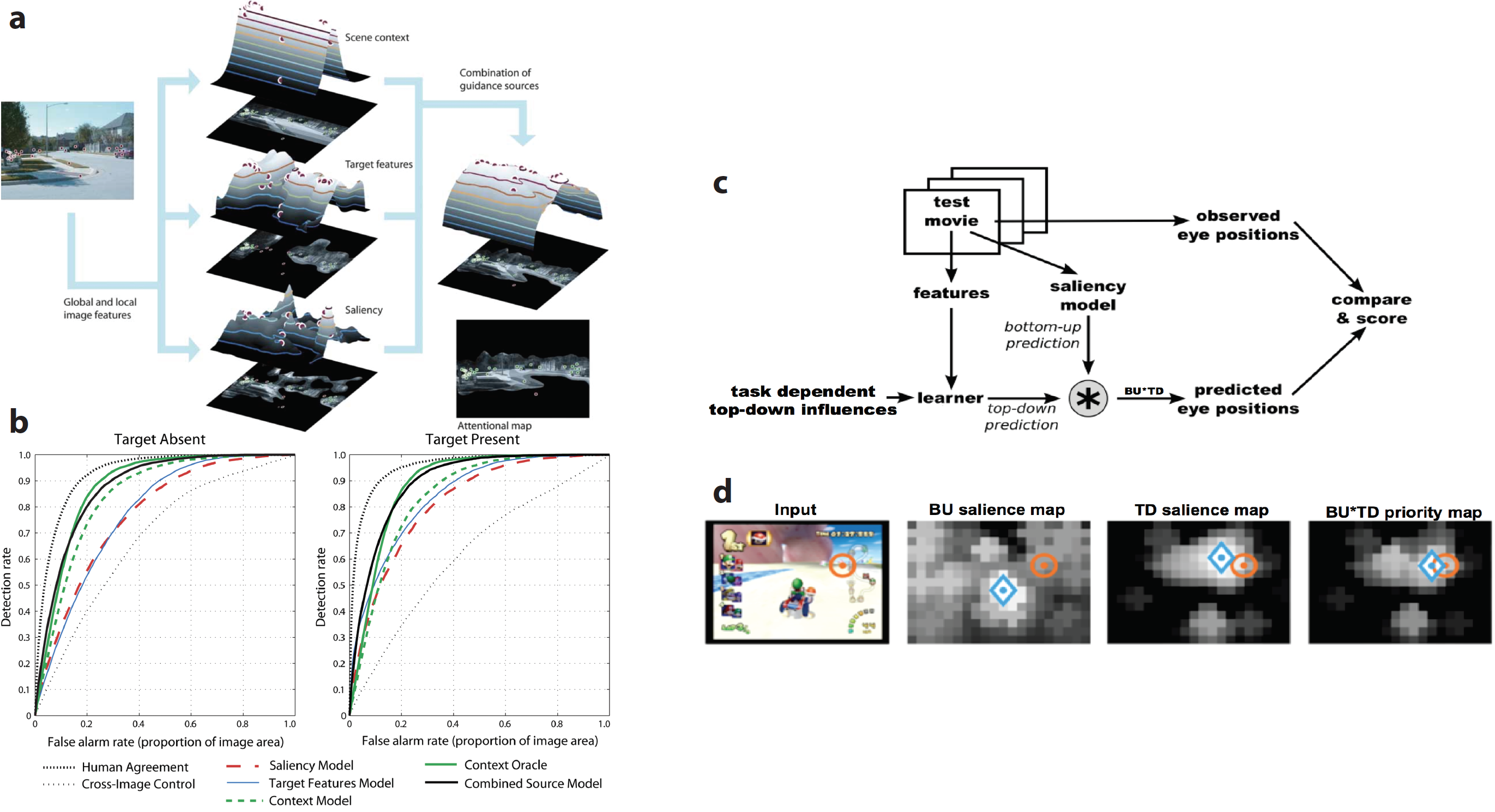}
\caption{Top-down models that involve spatial modulation. {\bf (a)} Model of Ehinger {\em at al.} (2009) where an
  attention map is informed by three guidance sources, given the task of finding people: (1) a spatial map that, based
  on the coarse scene structure, provides a spatial prior on where humans might appear in the given scene (e.g., they
  might appear on sidewalks); (2) A map that indicates where visual features in the image that resemble the features of
  the desired targets are observed; (3) a bottom-up saliency map. {\bf (b)} The combined-source model performs best and
  significantly better than any of the three component models taken alone, and also performs better than an empirical
  context oracle (where a set of humans manually indicated where humans might appear in the given scenes). {\bf (c)} A
  model that learns top-down priors from human gaze behavior while engaged in complex naturalistic tasks, such as
  driving (Peters \& Itti, 2007). A task-dependent learner component builds, during a training phase, associations
  between distinct coarse types of scenes and observed eye movements (e.g., drivers tend to look to the left when the
  road turns left). During testing, exposure to similar scenes gives rise to a top-down salience map, which is combined
  with a standard bottom-up salience map to give rise to the final (BU*TD) priority map that guides attention. {\bf (d)}
  Example results from the model of (c) applied to a driving video game. Blue diamonds represent the peak location in
  each map and orange circles represent the current eye position of the human driver. Here the bottom-up (BU) salience
  map considers that the main character is the most interesting scene element, but, as more correctly predicted by the
  top-down (TD) map, the driver is looking into the road's turn and on the horizon line. From Peters \& Itti (2007).}
\label{FIGtdmap}
\end{figure}

In a recent model, Ehinger {\em et al.}~\shortcite{Ehinger_etal09} investigated whether a model that, in addition to a
bottom-up saliency map, learns spatial priors about where people may appear and a feature prior about what people may
look like, would better predict gaze patterns of humans searching for people.  Indeed, several earlier studies had
suggested that bottom-up models, while widely demonstrated to correlate with human fixations during free viewing, may
not well predict fixations of participants once they are given a top-down task, for example a search task
\cite{Zelinsky_etal06,Foulsham_Underwood07,Henderson_etal07,Einhauser_etal08}. One may argue, however, especially in the
light of our above discussion of whether pure bottom-up salience is a valid concept, that in these experiments the
models were at an unfair disadvantage: Human participants had been provided with some information which had not been
communicated to models (e.g., search for a specific target, shown for 1 second before the search \cite{Zelinsky_etal06};
search of objects in a category or for a specific object \cite{Foulsham_Underwood07}; search for the small bullseye
pattern or for a local higher-contrast region \cite{Einhauser_etal08}; or count people \cite{Henderson_etal07}). Ehinger
{\em et al.} addressed this by proposing a model that combines three sources of information (Figure~\ref{FIGtdmap}.a):
First, a ``scene context'' map was derived from learning the associations between holistic or global scene features
(coarsely capturing the gist of the scene \cite{Torralba03}) and the locations where humans appeared in scenes with
given holistic features (trained over 1880 example images). This map, which is of central interest to this section of
our chapter, thus learned the typical locations where humans were expected to appear in different views of street
scenes. This learning step produces a prior on locations that can be used to filter out salient responses in locations
that are highly unlikely to contain the target (e.g., in the sky, assuming that no human was seen flying in the training
dataset). Second, a person detector was run in a sliding window manner over the entire image, creating a ``target
features'' map that highlighted locations that closely look like humans. This provides an alternative to learning
feature gains as discussed above; instead, an object detector algorithm is trained for the desired type of target. While
possibly more efficient than gain modulation, this approach suffers from lower biological plausibility. (See
\cite{Rao_etal02,Orabona_etal05} for related models).  Third and finally, a standard bottom-up ``saliency map'' provided
additional candidate locations (also see \cite{Oliva_etal03} for earlier related work, integrating only saliency and
scene context). Ehinger {\em et al.} found that the model which combined all three maps outperformed any of the three
component models taken alone (Figure~\ref{FIGtdmap}.b).

In a related model, Peters \& Itti \cite{Peters_Itti07cvpr} also used a combination of bottom-up saliency maps and
top-down spatial maps derived from the holistic gist of the scene, but their top-down maps were directly learned from
eye movements of human observers, playing the same 3D video games as would be used for testing (games included driving,
exploration, flight combat, etc.; note that since players control the game's virtual camera viewpoint, each run of such
game gives rise to a unique set of viewpoints and of generated scenes).  The bottom-up component of this model is based
on the Itti-Koch saliency model \cite{Itti_etal98pami}, which predicts interesting locations based on low-level visual
features such as luminance contrast, color contrast, orientation, and motion. The top-down component is based on the
idea of ``gist,'' which in psychophysical terms is the ability of people to roughly describe the type and overall layout
of an image after only a very brief presentation \cite{Li_etal02}, and to use this information to guide subsequent
target searches \cite{Torralba03}. This model (Figure~\ref{FIGtdmap}.c) decomposes each video frame into a low-level
image signature intended to capture some of the properties of ``gist'' \cite{Siagian_Itti07pami}, and learns to pair the
low-level signatures from a series of video clips with the corresponding eye positions; once trained, it generates
predicted gaze density maps from the gist signatures of previously unseen video frames. To test these bottom-up and
top-down components, we compared their predicted gaze density maps with the actual eye positions recorded while people
interactively played video games (Figure~\ref{FIGtdmap}.d).

\nocite{Peters_Itti07cvpr}

\subsection{More complex top-down models} \label{SECtdcomplex}

Many top-down models have been proposed which include higher degrees of cognitive scene understanding. Already in the
late 1990's several models included a top-down component that decided where to look next based on what had been observed
so far (e.g., \cite{Rybak_etal98,Schill_etal01}; also see \cite{Itti_Koch01nrn} for review). In robotics, the notion of
combining or alternating between different behaviors (such as exploration versus search, or bottom-up versus top-down)
has also led to several successful models \cite{Sprague_Ballard03,Forssen_etal08,Burattini_etal10,Xu_etal10}. More
recently, and our focus here, probabilistic inference and reasoning techniques, very popular in computer vision, have
started to be used in attention models.

In many recent models, the saliency map of bottom-up models is conserved as a data-driven source of information for an
overarching top-down system (more complicated than the feature or spatial biasing described above). For example,
Boccignone \& Ferraro \shortcite{boccignone2004modelling} develop an overt attention system where the top-down component
is a random walker that follows an information foraging strategy over a bottom-up saliency map. They demonstrate
simulated gaze patterns that better match human distributions \cite{Tatler_etal11}. Interesting related models have been
proposed where bottom-up and top-down attention interact through object recognition \cite{Ban_etal10,Lee_etal11}, or by
formulating a task as a classification problem with missing features, with top-down attention then providing a choice
process over the missing features \cite{Hansen_etal11}.

\begin{figure}[htb]
\includegraphics[width=\linewidth]{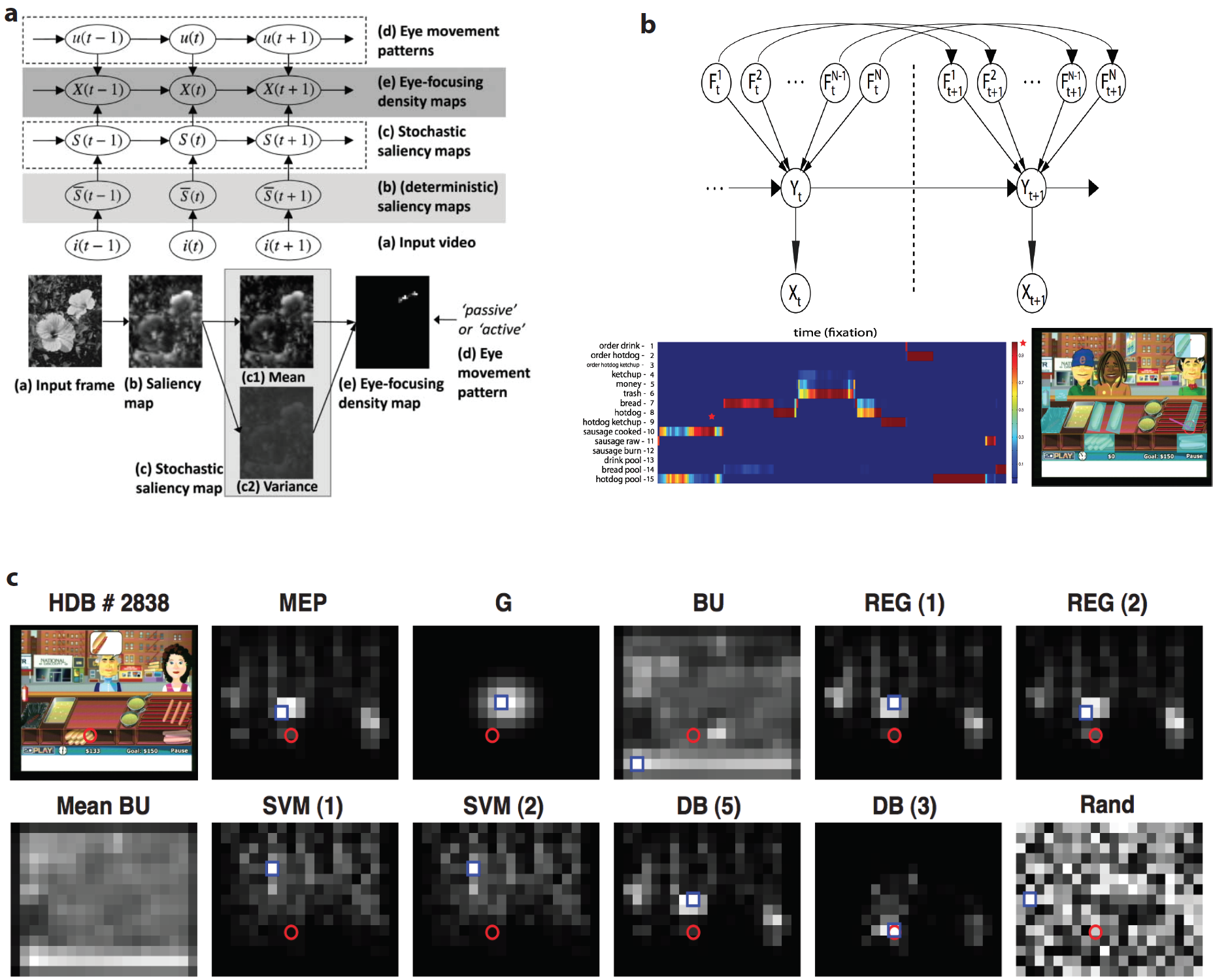}
\caption{Examples of recent more complex top-down models. {\bf (a)} Model of Akamine {\em et al.} (2012) which combines
  bottom-up saliency influences and top-down active/passive state influences over space and time using a dynamic
  Bayesian network. Although the top-down state is quite simple in this model (active vs.\ passive), the proposed
  mathematical framework could easily extend to more complex top-down influences. {\bf (b)} Graphical representation of
  the DBNs approach of Borji {\em et al.} (2012 AAAI) unrolled over two time-slices. $X_{t}$ is the current saccade
  position, $Y_{t}$ is the currently attended object, and $F^{i}_{t}$ is the function that describe object $i$ at the
  current scene. All variables are discrete. It also shows a time series plot of probability of objects being attended
  and a sample frame with tagged objects and eye fixation overlaid. {\bf (c)} Sample predicted saccade maps of the DBN
  model (shown in b). Each red circle indicates the observer’s eye position superimposed with each map’s peak location
  (blue squares). Smaller distance indicates better prediction.  Images from top-left to bottom-right are: a sample
  frame from the hot-dog bush game where the player has to serve customers food and drink, MEP stands for the mean eye
  position over all frames during the game play, G is just a trivial Gaussian map at the image center, BU is the
  bottom-up saliency map of the Itti model, REG(1) is a regression model which maps the previous attended object to the
  current attended object and fixation location, REG(2) is similar to REG(1) but the input vector consists of the
  available objects at the scene augmented with the previously attended object, SVM(1) and SVM(1) correspond to REG(1)
  and REG(2) but using an SVM classifier, Mean BU is the average BU map showing which regions are salient throughout the
  game course, Similarly DBN(1) and DBN(2) correspond to REG(1) and REG(2) meaning that in DBN(1) network slice consists
  of just one node for previously attended object while in DBN(2) each network slice consists of the previously attended
  object as well information of the previous objects in the scene, and finally Rand is a white noise random map.}
\label{FIGtdcomplex}
\end{figure}

Of growing recent interest is the use of probabilistic reasoning and graphical models to explore how several sources of
bottom-up and top-down information may combine in a Bayesian-optimal manner. For example, the model of Akamine {\em et
  al.} \shortcite{Akamine_etal12} (also see \cite{Kimura_etal08}), which employs probabilistic graphical modeling
techniques and considers the following factors, interacting in a dynamic Bayesian network (Figure~\ref{FIGtdcomplex}.a):
On the one hand, input video frames give rise to deterministic saliency maps. These are converted into stochastic
saliency maps via a random process that affects the shape of salient blobs over time (e.g., dynamic Markov random field
\cite{Kimura_etal08}). An eye focusing map is then created which highlights maxima in the stochastic saliency map,
additionally integrating top-down influences from an eye movement pattern (a stochastic selection between passive and
active state with a learned transition probability matrix). The authors use a particle filter with Markov chain
Monte-Carlo (MCMC) sampling to estimate the parameters; this technique often used in machine learning allows for fast
and efficient estimation of unknown probability density functions. Although the top-down component is quite simple in
this version of the model, it is easy to see how more sophisticated top-down and contextual influences could be
integrated into the dynamic Bayesian network framework of Kimura {\em et al.}  Several additional recent related models
using graphical models have been proposed (e.g., \cite{Chikkerur_etal10}).

Although few have been implemented as fully computational models, several efforts have started to develop models that
perform reasoning over objects or other scene elements to make a cognitive decision of where to look next
\cite{Navalpakkam_Itti05vr,Yu_etal08,Beuter_etal09,Yu_etal12}.

A a recent example, using probabilistic reasoning and inference tools, Borji~{\em et al.}~\cite{Borji_etal12aaai}
introduced a framework to model top-down overt visual attention based on reasoning, in a task-dependent manner, about
objects present in the scene and about previous eye movements.  They designed a Dynamic Bayesian Network (DBN) that
infers probability distributions over attended objects and spatial locations directly from observed data. Two basic
concepts in this model are 1) taking advantage of the sequence structure of tasks, which allows to predict the future
fixations from past fixations and knowledge about objects present in the scene. Graphical models have indeed been very
successful in the past to model sequences with applications in different domains, including biology, time series
modeling, and video processing, and 2) computing attention at the object level. Since objects are essential building
blocks in scenes, it is reasonable to assume that humans have instantaneous access to task-driven object-level variables
(as opposed to only gist-like, scene-global, representations).  Briefly, the model works by defining a Bayesian network
over object variables that matter for the task. For example, in a video game where one runs a hot-dog stand and has to
serve multiple hungry customers while managing the grill, those include raw sausages, cooked sausages, buns, ketchup,
etc.\~(Figure~\ref{FIGtdcomplex}.b). Then, existing objects in the scene, as well as the previous attended object,
provide evidence toward the next attended object (Figure~\ref{FIGtdcomplex}.b). The model also allows to read out which
spatial location will be attended, thus allowing one to verify its accuracy against the next actual fixation of the
human player. The parameters of the network are learned directly from training data in the same form as the test data
(human players playing the game). This object-based model was significantly more predictive of eye fixations compared to
simpler classifier-based models, also developed by the same authors, that map a signature of a scene to eye positions,
several state-of-the-art bottom-up saliency models, as well as brute-force algorithms such as mean eye position
(Figure~\ref{FIGtdcomplex}.c). This points toward the efficacy of this class of models for modeling spatio-temporal
visual data in presence of a task and hence a promising direction for future. Probabilistic inference in this model is
performed over object-related functions which are fed from manual annotations of objects in video scenes or by
state-of-the-art object detection models.  (Also see \cite{Sun_Fisher03,Sun_etal08} for models that consider objects,
although they do not reason about object identities and task-dependent roles).

\nocite{Borji_etal12aaai}

Finally, several computational models have started to explore making predictions that go beyond simply the next attended
location. For example, Peters \& Itti \cite{Peters_Itti08nips} developed a model that monitors in an online manner video
frames and eye gaze of humans engaged in 3D video games, computing instantaneous measures of how well correlated the eye
is with saliency predictions and with gist-based top-down predictions. They then learn to detect specific patterns in
these instantaneous measures, which allows them to predict~---~up to several seconds in in advance~---~when players are
about to fire a missile in a flight combat game, or to shift gears in a driving game. This model hence in essence
estimates the intentions and predicts the future actions of the player.  A related recent model was proposed by Doshi \&
Trivedi \cite{Doshi_Trivedi10} for active vehicle safety and driver monitoring. The system both computes bottom-up and
top-down saliency maps from a video feed of the driver's view, and monitors the eye movements of the driver to better
predict driver attention and gaze by estimating online the cognitive state and level of distraction of the driver. Using
similar principles and adding pattern classification techniques, Tseng {\em et al.} \shortcite{tseng2009quantifying}
have recently introduced a model that uses machine learning to classify, from features collected at the point of gaze
over a few minutes of television viewing, control subjects from patients with disorders that affect the attention and
oculomotor systems. The model has been successfully applied to elderly subjects (classifying patients with Parkinson's
disease vs.\ controls) as well as children (classifying children with Attention Deficit Hyperactivity Disorder
vs.\ Fetal Alcohol Spectrum Disorder vs.\ controls well above chance). These recent efforts suggest that eye movement
patterns in complex scenes do contain~---~like a drop of saliva~---~latent individual biomarkers, which the latest
attention modeling and pattern classification techniques are now beginning to reliably decode.

\section{Discussion and outlook} \label{SECdisc}

Our review shows that tremendous progress has been made in modeling both bottom-up and top-down aspects of attention
computationally. Tens of new models have been developed, each bringing new insight into the question of what makes some
stimuli more important to visual observers than other stimuli.

Our quantitative comparison of many existing computational models on three standard datasets (Figure~\ref{FIGauc})
prompts at least two reactions: First, it is encouraging to see that several models perform significantly better than
trivial models (e.g., a central Gaussian blob) or than older models (e.g., Itti-CIO2 model in
Figure~\ref{FIGauc}). Second, however, it is surprising that the ranking of model scores is quite substantially
different from the chronological order in which models were published. Indeed, one would typically expect that for a new
model to be recognized, it should demonstrate superior performance compared to the state of the art, and actually often
this is the case~---~just using possibly different datasets, scoring metrics, etc. Thus one important conclusion of our
study is that carrying out standardized evaluations is important to ensure that the field keeps moving forward (see
\cite{BorjiWeb} for a web-based effort in this direction).

Another important aspect of model evaluation is that currently almost all model comparisons and scoring are based on
average performance over a dataset of images or video clips, where often the dataset has been hand-picked and may
contain significant biases \cite{TorralbaEfros}. It may be more fruitful in the future to focus scoring on the most
dramatic mistakes a model might make, or the worst-case disagreement between model and human observers. Indeed, average
measures can easily be dominated by trivial cases if those happen often (e.g., we discussed earlier the notion of center
bias and how a majority of saccades which humans make are aimed towards the centers of images), and models may be
developed which perform well in these cases but miss conceptually important understanding of how attention may operate
in the minority of non-trivial cases. In addition, departure from average performance measures may provide richer
information about which aspects of attention are better captured by a given model (e.g., some models may perform better
on some sub-categories or even instances of images than others).

As we described more models, and in particular started moving from bottom-up models to those which include top-down
biases, the question arose of whether purely bottom-up models are indeed relevant to real life. In other words, is there
such a state of human cognition where a default or unbiased form of salience may be computed and may guide gaze. Many
experiments have assumed that free viewing, just telling observers to ``watch and enjoy'' stimuli presented to them,
might be an acceptable approximation to this canonical unbiased state. However, it is trivially clear from introspection
that cognition is not turned off during free viewing, and that what we look at in one instant triggers a range of
memories, emotions, desires, cognitive inferences, etc.\ which all will ultimately influence where we look next. In this
regard, it has been recently suggested that maybe only the initial volley of activity through visual cortex following
stimulus onset may represent such canonical bottom-up saliency representation \cite{Theeuwes10}. If such is the case,
then maybe comparing model predictions to sometimes rather long sequences of eye movements may be not be the best
measure of how well a model captures this initial purely bottom-up attention.

Models where top-down influences serve to bias the bottom-up processing stages have also blurred the line between
bottom-up and top-down. In fact, this is an important reminder that bottom-up and top-down influences are not mutually
exclusive and do not sum to give rise to attention control \cite{Awh_etal12}. Instead, bottom-up and top-down often
agree: The actor cognitively identified as the protagonist in a video clip may also move in such ways that he is the
most salient. In fact, today's bottom-up may be nothing more than our former generations' top-down. Indeed, some
bottom-up models have successfully integrated high-level features such as human face detectors into their palette of
feature maps \cite{Judd_etal09}, which blurs again the line between bottom-up and top-down (these features are computed
in a bottom-up manner from the image, but their very presence in a model is based on top-down knowledge that humans do
strongly tend to look at faces in images \cite{Cerf_etal09}).

When there is a task, top-down influences on attention are often believed to dominate, though this remains controversial
and depends both on the task and on the quantification method used
\cite{Zelinsky_etal06,Foulsham_Underwood07,Henderson_etal07,Einhauser_etal08,greene2012reconsidering}. In human vision,
we should not forget the following: Purely top-down attention (i.e., making a purely volitional eye movement unrelated
to any visual stimulus) is not a generally viable model, except maybe in blind persons. Others may make pure top-down
eye movements from time to time, but certainly not always~---~that is, no matter how strongly one believes that top-down
influences dominate, in the end controlling visual attention is a visually-guided behavior, and, as such, it is
dependent on visual stimuli. This is important for future modeling efforts, as they attempt to tackle more complex tasks
and situations, such as making a sandwich (Figure~\ref{FIGtd}.d): A person may indeed look at the jar of jam because it
is the next required object for the task (top-down guidance towards the jam). However, how did that person know where
the jam is? In most cases, some bottom-up analysis (maybe in the past) must have provided that information (except maybe
if the person was told where the jam is). Thus, modeling human behavior in complex tasks will likely require very
careful control over the experimental setup, so that human participants are not given more information or additional
priors that are not communicated to models (e.g., let a person look around before the task begins; see \cite{Foulsham12}
for recent relevant data). This consideration echoes our earlier remark about making fair comparisons: If a model is not
given the same information as a human participant (e.g., the model is not biased towards a search target or is not
allowed to explore a scene before the task begins), likely the model will not perform as well, but we will also learn
very little from such an experiment.

Another important challenge for models briefly mentioned above is dealing with sequence in eye movement data (i.e.,
scanpath) and with how to capture temporality in saccades. When comparing models, a model might be favored not only if
it can predict exact saccade locations, but also their ordering and their individual times of occurrence. In free
viewing, in spite of past efforts \cite{Privitera_Stark00}, it is still not clear whether such sequential
information is a strong factor of attention control and to what extent it depends on the subject or the asked question
\cite{Yarbus67}. Despite this, recently some researchers (e.g., \cite{Wang_etal11}) have tried to develop models and
scores to explain sequences of saccades. As opposed to free viewing, it seems that there is much more temporal
information in saccades in presence of a task. For instance, assume an observer is viewing videos of two different tasks
such as sandwich making or driving. It probably should not be very difficult to decode the task just from the sequential
pattern of eye movements.  This means that the task governs sequence of saccades when there is a task. In free-viewing,
however, when subjects are asked to watch a static scene freely there might not be a unique instruction making them to
saccade sequentially to certain places. Even if subjects are asked to watch the scene under different questions,
chances are that sequence may not help to decode the task \cite{greene2012reconsidering}.

Our survey shows that the remaining gap between man and machine seems to a large extent to be in 3D+time scene
understanding, which includes reconstruction of the 3D geometry of the scene, understanding temporal sequences of
events, simulation and extrapolation of physics over time in that 3D environment (e.g., to extrapolate the trajectory of
a ball as in Figure~\ref{FIGtd}.c), and so on. This requires some degree of machine vision and scene understanding which
is not yet solved in the general case. This means that future computational models of attention will need to bring to
bear sophisticated machine vision algorithms for scene understanding, to provide the necessary parsing of visual inputs
into tokens that can be reasoned upon and prioritized by attention.

\subsubsection*{Acknowledgements} Supported by the National Science
Foundation (grant numbers BCS-0827764 and CMMI-1235539), the Army Research Office (W911NF-11-1-0046 and 62221-NS), the
U.S. Army (W81XWH-10-2-0076), and Google. The authors affirm that the views expressed herein are solely their own, and
do not represent the views of the United States government or any agency thereof.

\bibliographystyle{authordate1}
\bibliography{/home/itti/bibliography/ilab,/home/itti/bibliography/master,reviewrefs}

\end{document}